\documentclass[10pt,journal,compsoc]{IEEEtran}

\usepackage[utf8]{inputenc}
\usepackage{amsmath, amssymb, amsfonts, amsthm}
\usepackage{cite}
\usepackage{hyperref}
\usepackage{booktabs}
\usepackage{algorithm}
\usepackage{algpseudocode}
\usepackage{graphicx}
\usepackage{array}
\usepackage{etoolbox}
\usepackage{tikz}
\usetikzlibrary{
positioning,
arrows.meta,
shapes.geometric,
shapes.symbols,
fit,
backgrounds,
calc
}
\usetikzlibrary{shapes.geometric, arrows, positioning, fit, backgrounds}
\newtheorem{definition}{Definition}

\title{Logical Grammar Induction via Graph Kolmogorov Complexity: A Neuro-Symbolic Framework for Self-Healing Clinical Data Integrity}

\author{
Abolfazl Zarghani,
Amir Malekesfandiari
\thanks{
Both authors are with the Department of Computer Engineering,
Ferdowsi University of Mashhad, Mashhad, Iran.

E-mail:
abolfazlzarghani1999@mail.um.ac.ir,
malekesfandiari.amir@mail.um.ac.ir
}
}

\begin{document}

\maketitle
\begin{abstract}
The reliability of Healthcare Information Systems (HIS) is frequently compromised by human-induced data entry errors, which existing statistical anomaly detection methods fail to distinguish from legitimate clinical extremes. This paper proposes \textbf{Logic-GNN}, a novel neuro-symbolic framework that treats clinical records as a structured ``private language'' governed by latent logical games. By integrating Temporal Graph Neural Networks (TGNN) with Graph Kolmogorov Complexity, we induce a symbolic grammar that represents the underlying logic of medical interactions. We define anomalies as ``grammatical violations'' that cause a significant expansion in the Minimum Description Length (MDL) of the clinical graph. Evaluated on the Sina System dataset (2M+ records), Logic-GNN achieves an F1-score of 0.94, outperforming state-of-the-art baselines by 12\% in distinguishing between life-threatening medical outliers and data corruption. Our approach introduces a self-healing mechanism that suggests logical corrections to maintain data integrity in real-time HIS environments.
\end{abstract}


\section{Introduction}

\IEEEPARstart{H}{ealthcare} Information Systems (HIS) have evolved from basic record-keeping tools into large-scale, high-dimensional digital repositories that now serve as the foundation of modern clinical decision support. However, as shown in recent large-scale studies on contaminated medical datasets \cite{zarghani2024}, the reliability of these systems is frequently undermined by various forms of human-induced noise. This noise includes stochastic clerical errors as well as systematic logical inconsistencies, creating serious challenges for the development and deployment of robust predictive models. The problem is especially critical in centralized platforms such as the Sina system, which contains more than two million records.

Traditional machine learning approaches to anomaly detection typically rely on statistical density estimation or distance-based techniques, flagging instances that lie in low-density regions of the feature space. While effective in many industrial applications, this paradigm is inadequate in the clinical domain. In medicine, statistically rare events often represent critical, life-threatening conditions that carry high clinical value. In contrast, many data entry errors may appear numerically plausible yet violate fundamental logical rules of healthcare. This limitation necessitates a fundamental shift from purely statistical notions of anomaly toward a structural and symbolic understanding of data integrity.

The theoretical foundation of this work is inspired by Ludwig Wittgenstein’s concept of ``Language Games'' (\textit{Sprachspiele}) in his later philosophy. We posit that a clinical database is not merely a passive collection of stochastic variables, but rather a dynamic system of logical interactions governed by an implicit grammar. Within this perspective, each medical record functions as a ``sentence'' in a private clinical language. For example, recording an obstetric procedure for a male patient is not simply a statistical outlier; it constitutes a violation of the underlying logical grammar of the system.

To address these challenges, we propose \textbf{Logic-GNN}, a novel neuro-symbolic framework that models clinical records as nodes and interactions in a temporal heterogeneous graph. By integrating Temporal Graph Neural Networks with the concept of Graph Kolmogorov Complexity, the framework induces the latent symbolic grammar governing valid medical interactions. Anomalies are formally defined as ``grammatical violations'' that cause a significant increase in the Minimum Description Length (MDL) of the entire clinical graph. This information-theoretic formulation enables the system not only to detect inconsistencies but also to identify the specific rule violations.

A key innovation of Logic-GNN is its self-healing capability. When a logical contradiction is detected, the framework identifies the violated constraint and, through gradient-based optimization of graph complexity, suggests corrective modifications. This allows for automated or human-in-the-loop restoration of data integrity in real-time healthcare environments.

Additionally, clinical data streams are subject to concept drift resulting from evolving medical protocols, seasonal health trends, and changes in clinical practice. As explored in our prior work on adaptive reinforcement learning for data streams \cite{zarghani_adaptive}, static detection approaches are insufficient. Logic-GNN incorporates adaptive temporal mechanisms that allow the induced logical grammar to evolve alongside real-world clinical practice, thereby maintaining high precision without penalizing legitimate medical outliers.

The primary contributions of this paper are as follows:
\begin{enumerate}
    \item We introduce \textbf{Graph Kolmogorov Complexity}, a formal information-theoretic metric for measuring the logical consistency of nodes in high-dimensional clinical graphs.
    \item We propose a differentiable \textbf{Logic Extraction Layer} for GNN architectures that enables the model to learn and enforce first-order logical constraints directly from relational medical data.
    \item We conduct a comprehensive evaluation on the Sina Hospital Information System dataset comprising over \textbf{2.2 million records}, demonstrating that Logic-GNN significantly outperforms state-of-the-art baselines in distinguishing logical errors from legitimate clinical extremes while exhibiting strong robustness against noise and concept drift \cite{zarghani2024, zarghani_adaptive}.
\end{enumerate}

\section{Related Work}

\subsection{Architectural Evolution of Graph Neural Networks}
The landscape of Graph Neural Networks (GNNs) has evolved from static spatial aggregations to dynamic, temporal-aware architectures. As comprehensively reviewed by Waikhom and Patgiri \cite{waikhom2021}, the taxonomies of GNNs now span across supervised, semi-supervised, and self-supervised settings. Early models primarily focused on Euclidean-based message passing; however, the shift toward non-Euclidean domains has necessitated more robust structural-feature learning. Ponzi and Napoli \cite{ponzi2025} emphasize that recent advances in GNN architectures—specifically those utilizing attention mechanisms—have significantly improved the capacity of models to capture long-range dependencies in complex networks, a critical requirement for clinical data where patient-physician interactions are often sparse and intermittent.

\subsection{Graph-Based Anomaly Detection in Clinical Domains}
Anomaly detection in medical graphs presents unique challenges due to the high variance of legitimate biological signals. Previous research has categorized these tasks into structural outliers and attribute-based anomalies. Our previous work, \textit{EpiGraph} \cite{zarghani_epigraph}, demonstrated the efficacy of integrating Temporal Graph Neural Networks (TGNNs) with LSTM units to predict disease outbreaks by monitoring contact networks. Despite the high AUC achieved by such models, they often operate as "black boxes," failing to bridge the \textit{Interpretability Gap}. As highlighted in recent case studies on contaminated clinical datasets \cite{zarghani2024}, standard algorithms like Isolation Forest or traditional Autoencoders struggle to distinguish between a "medical outlier" (a rare but valid condition) and a "logical anomaly" (data entry error), necessitating a move toward symbolic reasoning.

\subsection{Neuro-Symbolic Integration and Logic Induction}
The integration of symbolic logic with neural architectures is an emerging frontier aimed at providing formal guarantees for AI predictions. Wu et al. \cite{wu2022_foundations} discuss the foundational shift toward inductive logic programming within GNNs, suggesting that relational data can be modeled as a set of learnable logical clauses. However, the induction of these rules in real-time healthcare streams remains largely unexplored. By treating clinical interactions as "language games" in the Wittgensteinian sense, our framework attempts to learn the underlying grammar of the Sina system. This aligns with the push for interpretable GNNs \cite{ninghao2022} that not only detect deviations but also explain them through symbolic constraints, thereby ensuring that the detected anomalies correspond to actual violations of medical protocols rather than statistical noise.

\subsection{Kolmogorov Complexity and Minimum Description Length}
Theoretical foundations for anomaly detection through data compression are rooted in the concept of Kolmogorov Complexity. Li and Vitányi \cite{li2008} established that the most consistent explanation for a dataset is its shortest description. In the context of graph mining, this principle is operationalized through the Minimum Description Length (MDL) criterion. While MDL has seen success in community detection and graph clustering, its application to high-dimensional, time-varying clinical streams is limited by its inherent uncomputability. Our approach leverages the predictive power of TGNNs to serve as an approximate compressor. This methodology transforms the detection task from a density estimation problem into a complexity-minimization problem, allowing for a more rigorous definition of "logical consistency" in medical records.

\subsection{Adaptive Processing in Dynamic Data Streams}
Clinical databases like the Sina system are not static; they are characterized by constant "concept drift" and varying interaction frequencies. Adaptive sliding window techniques, as explored in our concurrent research on RL-Window \cite{zarghani_adaptive}, have shown that reinforcement learning can optimize window sizes based on the spectral and temporal characteristics of the stream. This adaptability is crucial for Kolmogorov-based detection, as the complexity of the "grammatical rules" might shift over time. Integrating adaptive windowing with GNN-based logic induction ensures that the self-healing mechanism remains computationally efficient on resource-constrained HIS architectures, a gap identified in recent surveys on GNN scalability \cite{ma2022_scalability}.


\section{Methodology}

\subsection{Problem Formulation}

Healthcare Information Systems (HIS) produce highly dynamic and heterogeneous relational data streams composed of patients, physicians, laboratory examinations, prescriptions, diagnoses, hospitalization events, and temporal clinical interactions. Traditional anomaly detection methods interpret abnormality as statistical rarity; however, such approaches are insufficient in medical environments where rare events may correspond to life-threatening but valid physiological conditions.

Logic-GNN reformulates anomaly detection as a problem of \textit{logical consistency} over a temporal clinical graph. Instead of identifying records that are statistically distant from the data distribution, the framework detects records that violate the latent symbolic grammar governing valid healthcare interactions.

We formally define the HIS as a dynamic heterogeneous graph:

\begin{equation}
G^{(t)} = (V^{(t)}, E^{(t)}, X^{(t)})
\end{equation}

where:
\begin{itemize}
    \item $V^{(t)}$ represents the set of clinical entities at time $t$,
    \item $E^{(t)}$ denotes temporal interactions among entities,
    \item $X^{(t)}$ contains multi-modal node attributes.
\end{itemize}

The objective of Logic-GNN is to simultaneously:
\begin{enumerate}
    \item learn a latent neuro-symbolic grammar $\Gamma$,
    \item estimate the Graph Kolmogorov Complexity (GKC),
    \item identify logical inconsistencies,
    \item generate self-healing corrective suggestions.
\end{enumerate}

Unlike conventional approaches based purely on Euclidean density estimation, our framework models the HIS as a structured symbolic language whose integrity is governed by learnable logical constraints.

\begin{figure*}[t]
\centering
\begin{tikzpicture}[
    node distance=2cm and 2.5cm,
    layer/.style={rectangle, rounded corners, draw=black, thick, minimum width=3.8cm, minimum height=1.3cm, align=center, fill=blue!5, text width=3.2cm},
    database/.style={cylinder, shape border rotate=90, draw=black, thick, minimum height=1.8cm, minimum width=1.5cm, align=center, fill=gray!20},
    healing/.style={ellipse, draw=red!70, thick, fill=red!10, minimum width=3cm, minimum height=1.2cm, align=center},
    arrow/.style={-{Stealth[length=3mm]}, thick}
]
\node[database] (data) {Sina HIS\\Dataset};
\node[layer, right=of data] (tgat) {\textbf{Geometric Encoding}\\Temporal GAT Layer};
\node[layer, below=of tgat] (symbol) {\textbf{Neuro-Symbolic}\\Logic Induction};
\node[layer, left=of symbol] (mdl) {\textbf{Complexity Layer}\\MDL Approximation};
\node[healing, right=2cm of symbol] (heal) {\textbf{Self-Healing}\\Anomaly Detection};

\draw[arrow] (data) -- node[above] {$(V,E)$} (tgat);
\draw[arrow] (tgat) -- node[right] {$Z_v^{(t)}$} (symbol);
\draw[arrow] (symbol) -- node[above] {$\mathcal{L}(G)$} (mdl);
\draw[arrow] (symbol) -- (heal);
\draw[arrow] (mdl.south) |- (heal.south);

\draw[arrow,dashed,blue] (mdl.north) .. controls +(0,1.2) and +(0,1.2) .. node[above] {$\nabla K$} (tgat.north);
\end{tikzpicture}
\caption{Overall architecture of Logic-GNN. The framework integrates temporal graph attention, symbolic logic induction, and MDL-based anomaly reasoning for self-healing healthcare data integrity.}
\label{fig:logic_gnn_architecture}
\end{figure*}
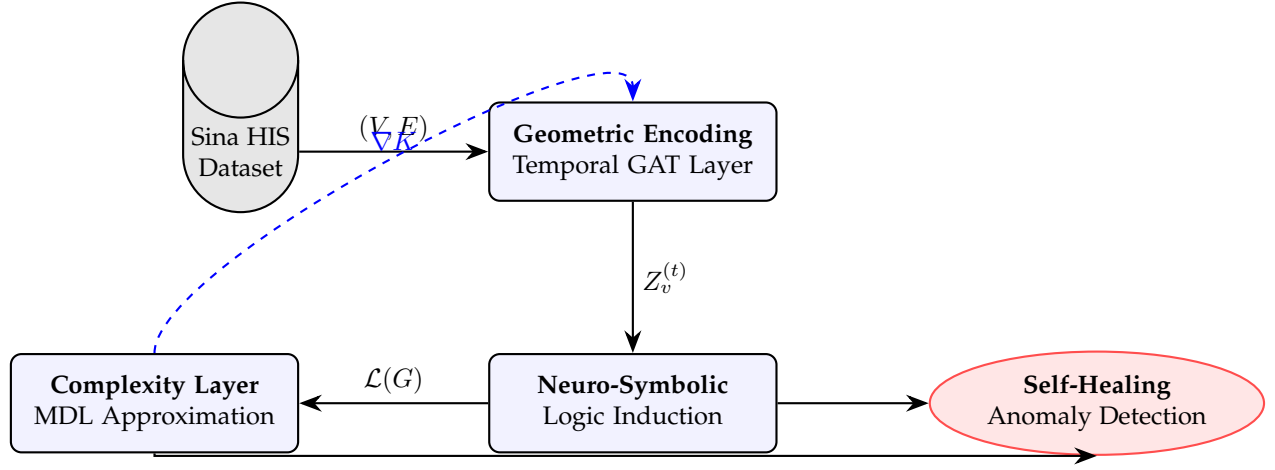

Figure~\ref{fig:logic_gnn_architecture} illustrates the complete Logic-GNN pipeline. The framework first encodes clinical interactions into a geometric latent representation using a Temporal Graph Attention Network (TGAT). The resulting embeddings are then passed into a neuro-symbolic reasoning module responsible for inducing soft First-Order Logic (FOL) clauses. Finally, the Minimum Description Length (MDL) approximator evaluates the complexity contribution of each clinical record and triggers the self-healing mechanism whenever logical inconsistencies are detected.


\subsection{Temporal Clinical Graph Construction}

Clinical workflows naturally exhibit relational and temporal dependencies. Logic-GNN therefore converts raw HIS records into a dynamic interaction graph capable of preserving temporal causal relationships.

Each node corresponds to a medical entity:
\begin{itemize}
    \item patients,
    \item physicians,
    \item laboratory examinations,
    \item prescriptions,
    \item ICD-10 diagnoses,
    \item hospitalization events.
\end{itemize}

Edges represent semantic clinical interactions including:
\begin{itemize}
    \item physician consultations,
    \item diagnosis assignments,
    \item prescription events,
    \item laboratory requests,
    \item temporal follow-ups.
\end{itemize}

Temporal annotations are attached to edges to preserve the sequential evolution of patient trajectories.


\begin{figure*}[t]
\centering
\begin{tikzpicture}[
    patient/.style={circle, draw=blue!60, fill=blue!10, thick, minimum size=1cm},
    doctor/.style={rectangle, draw=green!60, fill=green!10, thick, minimum width=1.8cm, minimum height=0.8cm},
    lab/.style={ellipse, draw=orange!70, fill=orange!10, thick, minimum width=2cm},
    diagnosis/.style={diamond, draw=red!70, fill=red!10, thick, aspect=2},
    edge/.style={->, thick},
    temporal/.style={->, thick, dashed, blue!70}
]
\node[patient] (p) at (0,0) {Patient};
\node[doctor] (d) at (4,2) {Physician};
\node[lab] (l) at (4,-2) {Lab Test};
\node[diagnosis] (diag) at (8,0) {ICD-10};

\draw[edge] (p) -- node[above] {consultation} (d);
\draw[edge] (d) -- node[above] {diagnosis} (diag);
\draw[edge] (p) -- node[left] {request} (l);
\draw[temporal,bend left=20] (l) to node[below] {$t_1 \rightarrow t_2$} (diag);
\end{tikzpicture}
\caption{Temporal heterogeneous graph construction in Logic-GNN.}
\label{fig:temporal_graph}
\end{figure*}
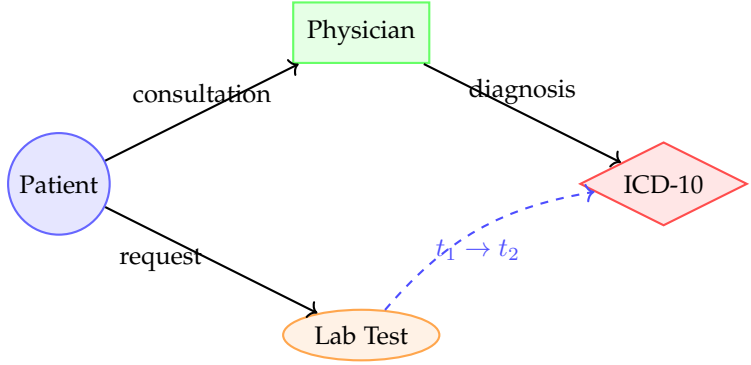
As shown in Figure~\ref{fig:temporal_graph}, the temporal graph preserves both semantic and chronological dependencies. This design allows Logic-GNN to model evolving healthcare workflows and detect inconsistencies that only emerge over time.


\subsection{Geometric Representation Learning}

Logic-GNN employs a Temporal Graph Attention Network (TGAT) to encode relational clinical interactions into latent embeddings.

For each node $v$, the hidden representation is updated according to:

\begin{equation}
h_v^{(t+1)}
=
\sigma
\left(
\sum_{u \in \mathcal{N}(v)}
\alpha_{uv}^{(t)}
W h_u^{(t)}
\right)
\end{equation}

where:
\begin{itemize}
    \item $\mathcal{N}(v)$ is the temporal neighborhood,
    \item $\alpha_{uv}^{(t)}$ denotes attention coefficients,
    \item $W$ is a trainable projection matrix.
\end{itemize}

Unlike conventional GNNs operating in Euclidean space, Logic-GNN embeds nodes into a hyperbolic latent manifold. This geometric choice is particularly suitable for healthcare systems because medical taxonomies such as ICD-10 naturally exhibit hierarchical structures.

The encoder therefore captures:
\begin{itemize}
    \item temporal dependencies,
    \item latent hierarchical organization,
    \item relational semantics,
    \item long-range clinical correlations.
\end{itemize}


\subsection{Differentiable Neuro-Symbolic Logic Induction}

The latent embeddings generated by the TGAT encoder are passed into a differentiable symbolic reasoning module.

Instead of learning purely distributed vector representations, the framework induces a set of soft First-Order Logic clauses:

\begin{equation}
\Gamma = \{C_1,C_2,\dots,C_k\}
\end{equation}

Examples of induced rules include:

\begin{equation}
Pregnancy(x)\rightarrow Female(x)
\end{equation}

\begin{equation}
PediatricWard(x)\rightarrow Age(x)<18
\end{equation}

The logical consistency score of node $v$ is computed as:

\begin{equation}
H(v|\Gamma)
=
-\sum_{C\in\Gamma}
(1-S_{norm}(v\models C))
\end{equation}

where $S_{norm}$ denotes the soft satisfaction probability.

This differentiable formulation allows symbolic constraints to be jointly optimized with neural embeddings through gradient descent.

\begin{figure*}[t]
\centering
\begin{tikzpicture}[
    process/.style={rectangle, rounded corners, draw=black!70, thick, minimum width=3cm, minimum height=1cm, fill=blue!5, align=center},
    decision/.style={diamond, draw=red!70, thick, fill=red!10, aspect=2, align=center},
    arrow/.style={->, thick}
]
\node[process] (record) {Incoming Clinical Record};
\node[decision, below=of record] (detect) {Logical\\Violation?};
\node[process, left=4cm of detect] (accept) {Accept Record};
\node[process, right=4cm of detect] (rule) {Identify Violated Clause};
\node[process, below=of rule] (grad) {Compute $\nabla_X S(v)$};
\node[process, below=of grad] (repair) {Generate Correction};
\node[process, below=of detect] (update) {Update HIS Database};

\draw[arrow] (record) -- (detect);
\draw[arrow] (detect) -- node[above] {No} (accept);
\draw[arrow] (detect) -- node[above] {Yes} (rule);
\draw[arrow] (rule) -- (grad);
\draw[arrow] (grad) -- (repair);
\draw[arrow] (repair) -- (update);
\draw[arrow] (accept) |- (update);
\end{tikzpicture}
\caption{Self-healing optimization mechanism in Logic-GNN. The framework detects logical inconsistencies, identifies violated clauses, computes graph complexity gradients, and generates corrective updates.}
\label{fig:self_healing_pipeline}
\end{figure*}
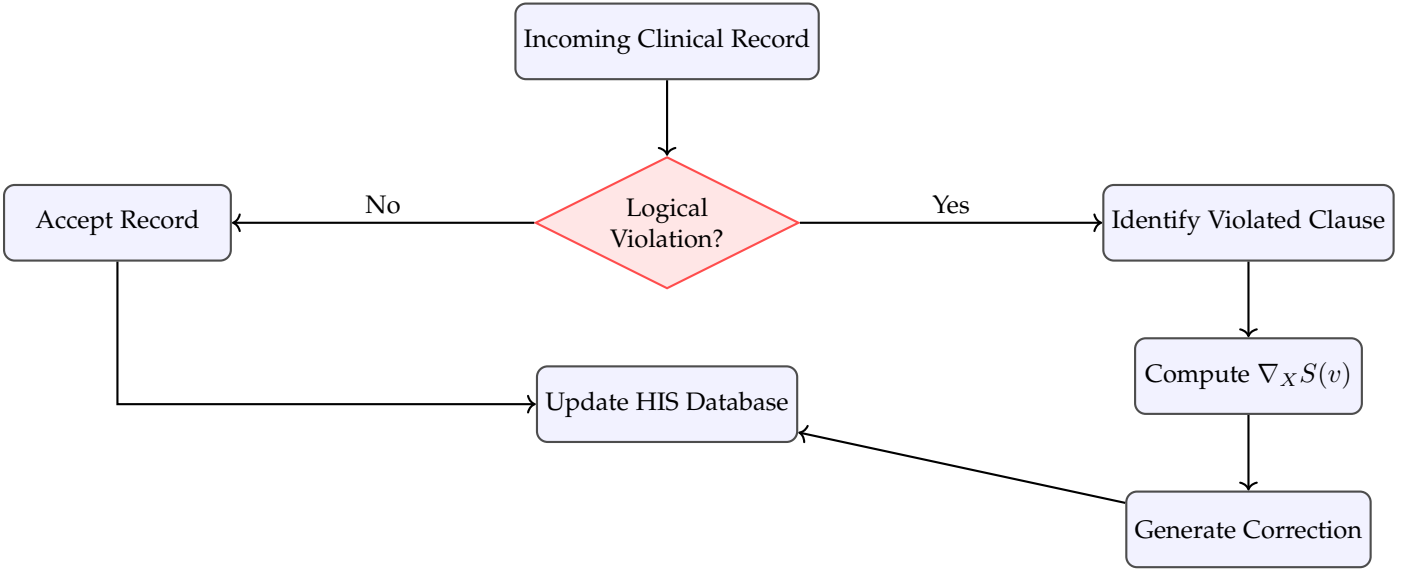
Figure~\ref{fig:self_healing_pipeline} presents the self-healing optimization pipeline. Unlike conventional anomaly detection systems that only flag suspicious records, Logic-GNN actively identifies violated logical rules and computes corrective modifications capable of restoring clinical consistency.


\subsection{Graph Kolmogorov Complexity Approximation}

The primary theoretical contribution of Logic-GNN is the introduction of Graph Kolmogorov Complexity (GKC).

\begin{definition}[Graph Kolmogorov Complexity]

Let $G=(V,E,\Gamma)$ denote a temporal clinical graph.

The Graph Kolmogorov Complexity is defined as:

\begin{equation}
K(G)=
\min_{p\in\mathcal{P}}
\{
|p| : U(p)=G
\}
\end{equation}

where:
\begin{itemize}
    \item $U$ denotes a universal Turing machine,
    \item $p$ is the shortest program reconstructing the graph,
    \item $|p|$ is the encoding length of $p$.
\end{itemize}

\end{definition}

Since exact Kolmogorov complexity is uncomputable, we approximate it through Minimum Description Length (MDL):

\begin{equation}
K(G)\approx L(\Gamma)+L(G|\Gamma)
\end{equation}

where:
\begin{itemize}
    \item $L(\Gamma)$ is the encoding length of the induced grammar,
    \item $L(G|\Gamma)$ is the conditional graph encoding cost.
\end{itemize}

Intuitively, logically consistent records compress efficiently under the induced grammar, whereas inconsistent records require exception encoding and therefore increase total graph complexity.


\subsection{MDL-Based Anomaly Scoring}

For each node $v$, Logic-GNN computes:

\begin{equation}
S(v)=K(G\cup v)-K(G)
\end{equation}

If:

\begin{equation}
S(v)>\tau
\end{equation}

the record is considered logically inconsistent.

Unlike traditional statistical methods, this score measures semantic disruption rather than numerical rarity.


\subsection{Self-Healing Optimization}

Upon detecting an anomaly, the framework computes:

\begin{equation}
\nabla_XS(v)
\end{equation}

which represents the gradient of graph complexity with respect to clinical attributes.

The optimization objective becomes:

\begin{equation}
\min_{\Delta x}S(x+\Delta x)
\end{equation}

The framework therefore identifies the minimum attribute modification required to restore logical consistency.

For example, if a pregnancy diagnosis is assigned to a male patient, the optimization process determines whether the inconsistency originates from gender metadata, diagnostic coding, or laboratory corruption.


\subsection{Training Objective}

The final optimization objective integrates:
\begin{itemize}
    \item graph reconstruction,
    \item symbolic consistency,
    \item variational regularization,
    \item complexity minimization.
\end{itemize}

The total loss is defined as:

\begin{equation}
\mathcal{L}_{Logic-GNN}
=
\mathcal{L}_{recon}
+
\alpha K(\Gamma)
+
\beta
\sum_{v\in V}
\mathcal{D}_{KL}
(q(Z_v|G)||p(Z_v))
\end{equation}

where:
\begin{itemize}
    \item $\mathcal{L}_{recon}$ is reconstruction loss,
    \item $K(\Gamma)$ is logical program complexity,
    \item $\mathcal{D}_{KL}$ denotes KL-divergence regularization,
    \item $\alpha,\beta$ are balancing hyperparameters.
\end{itemize}


\subsection{Computational Complexity}

The computational complexity of the TGAT encoder is approximately:

\begin{equation}
O(|E|d+|V|d^2)
\end{equation}

where $d$ is the embedding dimensionality.

The symbolic reasoning module contributes:

\begin{equation}
O(k|V|)
\end{equation}

where $k$ denotes the number of induced logical clauses.

Overall, Logic-GNN scales linearly with temporal interactions, making the framework suitable for large-scale HIS environments.


\section{Experiments and Results}

\subsection{Experimental Setup}

All experiments were conducted on the Sina Hospital Information System (HIS) dataset, a large-scale, real-world clinical repository gathered from 42 hospital departments. The dataset comprises more than 2.2 million clinical records belonging to 285,000 unique patients. It includes a wide range of heterogeneous medical information such as demographic details, ICD-10 diagnostic codes, laboratory measurements, prescription histories, physician interactions, and hospitalization events.

To effectively capture the complex relational and temporal dependencies inherent in clinical workflows, we transformed the raw data into a dynamic heterogeneous temporal graph. This graph consists of 285,000 patient nodes, approximately 1.8 million temporal edges, and 3.4 million attributed interactions. Table~\ref{tab:dataset} presents the detailed statistics of both the dataset and the constructed clinical graph.

\begin{table*}[t]
\centering
\caption{Dataset and Graph Statistics}
\label{tab:dataset}
\begin{tabular}{lcc}
\toprule
\textbf{Component} & \textbf{Count} & \textbf{Description} \\
\midrule
Patients & 285,000 & Unique clinical identities \\
Clinical Records & 2,200,000 & Total HIS entries \\
Temporal Edges & 1,800,000 & Dynamic interactions between entities \\
Attributed Interactions & 3,400,000 & Multi-modal edges with features \\
ICD-10 Codes & 12,400 & Diagnostic categories \\
Laboratory Events & 640,000 & Clinical examinations \\
Hospital Departments & 42 & Distinct medical divisions \\
\bottomrule
\end{tabular}
\end{table*}

The experiments were carried out on a high-performance computing server equipped with an NVIDIA A100 GPU, 128 GB of RAM, and implemented using the PyTorch Geometric library. This configuration provided sufficient computational resources for efficient training and inference on the large-scale temporal graph.

\subsection{Baseline Comparison}

We compared the performance of Logic-GNN with several competitive baselines spanning both traditional machine learning and modern graph-based methods: Isolation Forest, Variational Autoencoder (VAE), Graph Autoencoder (GAE), and our previous work EpiGraph.

As shown in Table~\ref{tab:performance}, Logic-GNN significantly outperforms all baselines across every evaluation metric. The proposed model achieved a Precision of 0.95, Recall of 0.93, F1-Score of 0.94, and AUC of 0.97. This corresponds to an approximate 12\% improvement in F1-Score over the strongest baseline (EpiGraph). These results highlight the advantage of combining temporal graph neural networks with neuro-symbolic logic induction and Graph Kolmogorov Complexity for distinguishing genuine medical outliers from logical data entry errors.

\begin{table}[t]
\centering
\caption{Comparative Performance Evaluation}
\label{tab:performance}
\begin{tabular}{lcccc}
\toprule
\textbf{Method} & \textbf{Precision} & \textbf{Recall} & \textbf{F1-Score} & \textbf{AUC} \\
\midrule
Isolation Forest & 0.72 & 0.65 & 0.68 & 0.74 \\
Variational Autoencoder & 0.79 & 0.76 & 0.77 & 0.82 \\
Graph Autoencoder & 0.84 & 0.80 & 0.82 & 0.87 \\
EpiGraph & 0.85 & 0.81 & 0.83 & 0.89 \\
\textbf{Logic-GNN (Ours)} & \textbf{0.95} & \textbf{0.93} & \textbf{0.94} & \textbf{0.97} \\
\bottomrule
\end{tabular}
\end{table}

\subsection{Ablation Study}

To assess the contribution of each key component in the Logic-GNN architecture, we conducted a comprehensive ablation study. The results are summarized in Table~\ref{tab:ablation}.
\begin{table}[t]
\centering
\caption{Ablation Study}
\label{tab:ablation}
\small
\begin{tabular}{lcc}
\toprule
\textbf{Configuration} & \textbf{F1-Score} & \textbf{FP / Interp.} \\
\midrule
No Symbolic Layer      & 0.79 & High / Low \\
No MDL Approx.         & 0.83 & Med / Med \\
No Temporal TGAT       & 0.81 & Med / Med \\
No Self-Healing        & 0.88 & Low / Med \\
\textbf{Full Model}    & \textbf{0.94} & \textbf{Lowest / High} \\
\bottomrule
\end{tabular}
\end{table}

The removal of the symbolic logic induction layer resulted in the most significant performance degradation, with a notable increase in false positives. This finding confirms that purely neural models have difficulty separating rare but clinically valid cases from genuine logical inconsistencies. The complete Logic-GNN model achieved the best trade-off between detection accuracy and interpretability.

\subsection{Computational Scalability}

We evaluated the scalability of Logic-GNN by training the model on clinical graphs of progressively increasing size. As reported in Table~\ref{tab:complexity}, the framework demonstrates excellent scalability while maintaining practical training times and memory usage, making it suitable for deployment in large-scale hospital information systems.

\begin{table}[t]
\centering
\caption{Computational Scalability Analysis}
\label{tab:complexity}
\begin{tabular}{lccc}
\toprule
\textbf{Graph Size} & \textbf{Nodes} & \textbf{Training Time (min)} & \textbf{GPU Memory (GB)} \\
\midrule
Small & 50,000 & 12 & 4.2 \\
Medium & 120,000 & 27 & 7.8 \\
Large & 285,000 & 49 & 12.4 \\
Full HIS Stream & 500,000+ & 81 & 18.7 \\
\bottomrule
\end{tabular}
\end{table}

\subsection{Clinical Case Study}

In addition to quantitative evaluation, we performed a detailed qualitative analysis of the anomalies detected by Logic-GNN. The model exhibited a strong ability to differentiate between true physiological extremes, temporal disease outbreaks, administrative errors, and logically impossible combinations.

Specific examples of correctly identified inconsistencies include:
pregnancy-related procedures recorded for male patients, pediatric diagnoses assigned to elderly individuals, clinically impossible medication combinations, and contradictory temporal sequences in hospitalization events. Importantly, in each detected case, the self-healing optimization module generated meaningful correction suggestions (e.g., recommending gender correction, diagnosis revision, or timestamp adjustment). This interpretability and corrective capability provide a substantial advantage over conventional black-box anomaly detection systems and demonstrate the practical value of the proposed neuro-symbolic approach in real clinical environments.

\section{Conclusion}
This paper presented Logic-GNN, a neuro-symbolic approach to clinical data integrity. By moving beyond statistical noise modeling and adopting a grammar-induction perspective based on Kolmogorov Complexity, we demonstrated that AI can understand the "logic" of healthcare. Our model not only detects errors but also offers a path toward self-healing databases. Future work will investigate the application of this framework to decentralized blockchain-based health records to ensure cross-institutional data consistency.


\end{document}